\documentclass[english]{IEEEtran}
\pdfoutput=1
\usepackage[T1]{fontenc}
\usepackage[latin9]{inputenc}
\usepackage{verbatim}
\usepackage{float}
\usepackage{amstext}
\usepackage{graphicx}

\makeatletter

\providecommand{\tabularnewline}{\\}
\floatstyle{ruled}
\newfloat{algorithm}{tbp}{loa}
\providecommand{\algorithmname}{Algorithm}
\floatname{algorithm}{\protect\algorithmname}

\newenvironment{lyxcode}
{\par\begin{list}{}{
\setlength{\rightmargin}{\leftmargin}
\setlength{\listparindent}{0pt}
\raggedright
\setlength{\itemsep}{0pt}
\setlength{\parsep}{0pt}
\normalfont\ttfamily}%
 \item[]}
{\end{list}}

\makeatother

\usepackage{babel}
\begin{document}

\title{Sequence Classification with Neural Conditional Random Fields}

\author{Myriam Abramson\\
Naval Research Laboratory\\
Washington, DC 20375\\
myriam.abramson@nrl.navy.mil}
\maketitle
\begin{abstract}
The proliferation of sensor devices monitoring human activity generates
voluminous amount of temporal sequences needing to be interpreted
and categorized. Moreover, complex behavior detection requires the
personalization of multi-sensor fusion algorithms. Conditional random
fields (CRFs) are commonly used in structured prediction tasks such
as part-of-speech tagging in natural language processing. Conditional
probabilities guide the choice of each tag/label in the sequence conflating
the structured prediction task with the sequence classification task
where different models provide different categorization of the same
sequence. The claim of this paper is that CRF models also provide
discriminative models to distinguish between types of sequence regardless
of the accuracy of the labels obtained if we calibrate the class membership
estimate of the sequence. We introduce and compare different neural
network based linear-chain CRFs and we present experiments on two
complex sequence classification and structured prediction tasks to
support this claim. \end{abstract}


\section{Introduction}

The proliferation of sensor devices monitoring human activity generates
voluminous amount of temporal sequences needing to be interpreted
and categorized. Moreover, complex behavior detection requires the
personalization of multi-sensor fusion algorithms. For example, stress
detection from physiological measurements can involve the fusion of
several variables such as pupil dilation, heart rate, and skin temperature.
In addition, it is the relative measurement to other states rather
than their absolute value that is more indicative of stress requiring
the monitoring of sequences of states and their transitions. %
Conditional random fields (CRFs) are commonly used in structured prediction
tasks such as part-of-speech tagging in natural language processing.
Conditional probabilities guide the choice of each tag/label in the
sequence conflating the structured prediction task with the sequence
classification task where different models could provide different
evaluations of the same sequence. The claim of this paper is that
CRF models also provide discriminative models to distinguish between
types of sequence regardless of the accuracy of the labels obtained,
provided that the class membership estimate of the sequence be calibrated.
In other words, the score obtained by the CRF model representing a
ranking of the sequence given the model has to correlate with the
empirical class membership probability \cite{zadrozny2002transforming}.
The intuition underlying this claim is that hard-to-detect complex
observation patterns might not provide an accurate labeling while
providing enough discriminatory evidence against other types of sequence.
CRFs are a very flexible way of modeling variable-length sequences
that can leverage from state-of-the-art discriminative learners. %
We present experiments in the hand-writing word recognition task and
in the Web analytics authentication task. 

This paper is organized as follows. Section \ref{sec:Related-Work}
describes related work in sequence classification. Section \ref{sec:Conditional-Random-Fields}
provides an overview of CRFs. Section \ref{sec:Methodology} presents
our methodology combining neural networks with CRFs. Section \ref{sec:Empirical-Evaluation}
presents our modeling and empirical evaluation of the hand-writing
word recognition task and of the Web analytics personalization task.
Section \ref{sec:Conclusion} concludes with discussion and future
work in this area.

\section{Related Work \label{sec:Related-Work}}

The problem that sequence classification addresses is introduced in
\cite{xing2010brief}, namely sequence classification is defined as
learning the function mapping a sequence $s$ to a class label $l$
from a set of labels $L.$ A distinction is made between predicting
sequence labels where the entire sequence is available and a sequence
of labels such as found in streaming data and addressed by structured
prediction methods. This latter problem is termed the strong classification
task. We argue in this paper that a stronger classification task might
be to identify a type of sequence from the classification of its temporal
components. Sequence classification methods include feature vectors,
distance-based methods and model-based methods like discriminative
$k$-Markov models or hidden Markov models (HMMs) if the sequence
labels are not observed at test time. 

Neural conditional random fields or \textit{neuroCRFs} have been investigated
in \cite{do2010neural} using deep neural networks showing the influence
of 2 layers vs. 1 layer of hidden nodes as well as the number of hidden
nodes in reducing the error rate for structured prediction tasks.
In this work, the outputs of the deep neural network compute the weights
of all the factors and leave the probabilistic framework of CRF intact. 

Similar to HMMs, Long Short Term Memory (LSTM) recurrent neural networks
(RNNs) \cite{DBLP:journals/corr/Graves13} learn a sequence of labels
from unsegmented data such as that found in handwriting or speech
recognition. This capability, called temporal classification, is distinguished
from framewise classification where the training data is a sequence
of pairwise input and output labels, suitable for supervised learning,
and where the length of the sequence is known. LSTM RNNs' architecture
consists of a hidden layer recurrent neural network with generative
capabilities and adapted for deep learning with skip connections between
hidden nodes at different levels. Unlike HMMs, there are no direct
connections between the output nodes of the neural network (i.e.,
the labels of the sequence), but there are indirect connections through
a prediction network from an output node to the next input. Consequently,
LSTM RNNs can do sequence labeling as well as sequence generation
through their predictive capability. 

In \cite{collins2002discriminative}, perceptrons were integrated
as discriminative learners in the probabilistic framework of CRFs
in the context of part-of-speech tagging. The Viterbi decoding algorithm
finds the best tagged sequence under the current weight parameters
of feature-tag pairs. As in the perceptron algorithm, weight updates
(0/1 loss) are triggered only when discrepancies occur. 

A distinction is made between discriminative and generative $k$-Markov
models in the sequence classification of simple symbolic sequences
\cite{yakhnenko2005discriminatively}. Generative models estimate
the probability distribution of features given a class from the training
data and base their classification decision on the joint probability
of the features and the class while discriminative models directly
estimate the conditional probability of a class given the features.
Similar to CRFs, discriminative $k$-Markov models maximize, using
gradient descent, two sets of parameters, namely, the probability
of a symbol $s_{i}$ given preceding symbols in a given model and
the joint probability of a sequence of symbols (assuming independence)
given a model. In addition, this work shows that a hybrid approach
initializing parameters with generative models can speed up convergence.

\section{Conditional Random Fields \label{sec:Conditional-Random-Fields}}

CRFs are a supervised method for structured prediction similar to
HMMs \cite{Rabiner89atutorial} while relaxing the independence assumption
of the observations and the Markov assumption \cite{lafferty:2001}.
The labels of the sequence must be provided at training time. CRFs
address the strong classification problem \cite{xing2010brief} of
predicting a sequence of labels but also take advantage of information
from the entire sequence to estimate the probability of the entire
sequence and therefore can also address the sequence classification
problem. We distinguish between weakly-supervised CRFs where the labels
are learned through an auxiliary classifier and strongly-supervised
CRFs where the labels are known without ambiguity \cite{abramson:socialcom}.
We describe below the derivation of CRFs from basic probabilistic
principles and restrict our discussion to linear-chain CRFs for the
classification of sequences. 

The factorization of Bayesian nets according to conditional independence
enables the tractable computation of the joint probability of a\label{eq:1-1}
collection of random variables $P(\bar{y})$ according to the structure
of a graphical model as follows. 

\begin{equation}
P(\bar{y})=\prod_{i}p(y_{i}|y_{i}^{p})\label{eq:1}
\end{equation}
where $y_{i}^{p}$ are the parents of $y_{i}$. However, it is sometimes
more natural to model a problem according to spatial or temporal proximity
of the nodes rather than their conditional independence. For example,
in a lattice-like graphical structure, the Markov blanket of a node
does not obey the spatial neighborhood properties of the graph as
expected \cite{murphy2012machine}. It is therefore more natural to
model such graphs as undirected graph models where the independence
of the nodes is determined only by the absence of a connecting edge.
It is possible to convert a directed graph to an undirected graph
by ``moralizing'' it (i.e. adding edges between nodes to indicate
implicit dependence). The edges of an undirected graph cannot be weighted
by conditional probabilities anymore but can be evaluated according
to the ``affinity'' of the nodes defined by a \textit{potential}
function or\textit{ factor $\varphi(x,y)$.} Those factors are parameterized
by a weight $\theta_{c}$ that can be learned from data using various
methods. According to the Hammersley-Clifford theorem, the joint probability
of the graph can then be obtained as follows:

\begin{equation}
P(\bar{y}|\bar{\theta})=\frac{1}{Z(\theta)}\prod_{c}\varphi(y_{c}|\theta_{c})\label{eq:2}
\end{equation}
where $Z(\theta$) is the partition function normalizing the product
of factors in order to obtain a probability distribution. 

\begin{figure}
\begin{centering}
\includegraphics[scale=0.29]{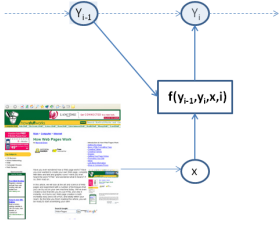}
\par\end{centering}

\caption{CRF feature function relating observations and labels. \label{fig:CRFs-feature-function}}
\end{figure}

CRFs leverage from the undirected graph modeling approach to model
the conditional distribution $P(\bar{y}|\bar{x})$ of a set of target
variables $\bar{y}$ and a set of observed variables $\bar{x}$ to
represent structured data. A factor represents the probability of
a target variable $y$ as a linear function of the weight parameters
$\theta$ and the observed input variables $\bar{x}$ which are not
necessarily independent. If the observed input variables are indicator
functions, $\phi(x,y),$ then $P(y|\bar{x},\bar{\theta})$ is defined
as follows:

\begin{equation}
P(y|\bar{x},\bar{\theta})=\frac{\exp\sum_{j}\theta_{j}\phi_{j}(\bar{x},y)}{\sum_{y'\in Y}Z(\bar{x},y')}\label{eq:3}
\end{equation}
where $Z(\bar{x},y')=\exp\sum_{j}\theta_{j}\phi_{j}(\bar{x},y')$.
The weight parameters $\theta$ of the factors are typically learned
using discriminative learning methods for the target variable $y$
as an alternative to the probabilistic likelihood estimation method
for computational efficiency reasons. There are two types of feature
functions in representing a sequence \cite{sutton2010introduction}:
(1) edge functions between two labels and (2) observation functions
relating $x$ and $y$. Generally, $f_{j}(y_{t-1},y_{t},\bar{x},t)$
represents a feature function combining both edge and observation
functions (Fig. \ref{fig:CRFs-feature-function}). The following are
examples of both types of feature functions in the handwriting recognition
domain:

\begin{flushleft}
{\footnotesize{}
\[
\phi_{j}(y_{i-1},y_{i})=\left\{ \begin{array}{cc}
1 & if\,\,y_{i}="i"\,\,and\\
 & y_{i-1}="n"\\
0 & otherwise
\end{array}\right.
\]
}
\par\end{flushleft}{\footnotesize \par}

{\footnotesize{}
\[
\phi_{j}(y_{i},x_{i})=\left\{ \begin{array}{cc}
1 & if\,\,pixel(x_{i})=1\,\,and\\
 & y_{i}="i"\\
0 & otherwise
\end{array}\right.
\]
}{\footnotesize \par}

The number of possible feature functions can be large but can be practically
restricted to those found in the training set. 

Generalizing to ``global'' factors over the entire sequence of observations
where $F_{j}(\bar{x},\bar{y})=\sum_{t}^{n}f_{j}(y_{t-1},y_{t},\bar{x},t)$
and where $t$ is the position in the sequence of length $n$, the
probability of the sequence $\bar{y}$ is then:

\begin{equation}
P(\bar{y}|\bar{x},\bar{\theta})=\frac{1}{Z(\bar{x},\bar{\theta})}\exp(\sum_{j}\theta_{j}F_{j}(\bar{x},\bar{y}))\label{eq:4-1}
\end{equation}
where $Z(\bar{x},\bar{\theta})=\sum_{\bar{y}\in nP_{Y}}\exp(\sum_{j}\theta_{j}F_{j}(\bar{x},\bar{y})).$
The normalization constant $Z$ is computationally intractable \cite{sutton2010introduction}
but does not need to be computed when predicting the labels in the
sequence given the weights of the feature functions or when evaluating
a sequence against one model. 

\section{Methodology \label{sec:Methodology}}

We address the problem of sequence learning with a potentially infinite
set of labels by learning several models. As in \cite{do2010neural},
we use the energy output of the output nodes, $E(x,y)$, before squashing
by the softmax function, as the score of the factors in the CRF. We
leverage from the neural network to discover non-linear feature functions
in the case of the multi-layer perceptrons (MLPs) and we compare and
contrast different neural network architectures. The Viterbi algorithm
\cite{Rabiner89atutorial,collins2002discriminative} guides the step-by-step
predictions to maximize the choice of each label $y$ with respect
to the entire sequence. $P(\bar{y}|\bar{x},\bar{\theta})$ is then
defined as follows:

\begin{equation}
P(\bar{y}|\bar{x},\bar{\theta})=\frac{\arg\max_{t_{1}...t_{n}}\prod_{t=1}^{n}P(y_{t}|y_{t-1},\bar{x}_{t},\bar{\theta})}{Z(\bar{x},\bar{\theta})}\label{eq:viterbi}
\end{equation}

Algorithm \ref{alg:Viterbi-algorithm-for-CRFs} describes the Viterbi
evaluation of a sequence of observations $\bar{x}$ delimited by START
and STOP tags combined with an approximate probabilistic evaluation
of the entire sequence given the model. $P(\bar{y}|\bar{x},\bar{\theta})$
can be approximated using a partition function, $Z(\bar{x},\bar{\theta})$,
that includes the maximum scores from the preceding step at each time
step rather than the sum of scores of all possible sequences. 

\begin{algorithm}[h]
input: model, $\bar{x}$ {\footnotesize{}//neural net and observation
sequence}{\footnotesize \par}

output: $\bar{y}$, $P(\bar{y}|\bar{x},\bar{\theta})$ {\footnotesize{}//label
sequence and probability}{\footnotesize \par}

\textbf{viterbi\_crfs} (model,$\bar{x}$)=

~~~t $\leftarrow$ 0

~~~$\alpha${[}t{]}$\leftarrow$ initialize ($y_{0},x_{0})$

\textbf{~~~while} t < length ($\bar{x}$)

~~~~~~t $\leftarrow$t + 1

~~~~~~$\alpha${[}t{]} $\leftarrow$ forward ($\bar{x}_{t}$,
$\alpha[t-1]$)

\textbf{~~~end}

~~~$y_{t}\leftarrow\arg\max_{y}\alpha[t]$

~\textbf{~~}$\bar{y}$ $\leftarrow$ backtrack ($y_{t}$,$\alpha$)

~~~score $\leftarrow$ $\max\alpha[t]$

~~~$P(\bar{y}|\bar{x},\bar{\theta})$ $\approx$ $\frac{exp(score)}{\sum_{y}\exp(\alpha_{y}[t])}$

\textbf{return} $\bar{y}$, score, $P(\bar{y}|\bar{x},\bar{\theta})$
\begin{lyxcode}

\end{lyxcode}
\caption{Viterbi algorithm for CRFs where forward and backtrack are functions
as in the Viterbi algorithm and where alphas contains the information
of all the possible outputs $y_{t}$ at each step $t$.\label{alg:Viterbi-algorithm-for-CRFs}}
\end{algorithm}

\begin{figure}
\begin{centering}
\includegraphics[scale=0.4]{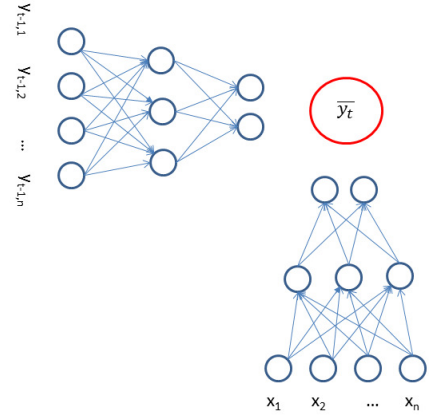}
\par\end{centering}

\caption{CRF-MLP - two MLPs combine to predict $y_{t}$ \label{fig:CRF-MLP}}
\end{figure}

\begin{figure}

\begin{centering}
\includegraphics[scale=0.4]{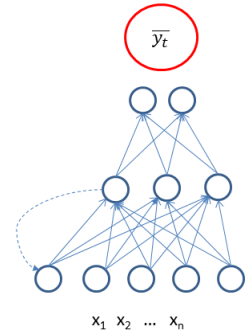}
\par\end{centering}

\caption{CRF-RNN - Elman network architecture where hidden node activations
from the previous time step are added to the current input\label{fig:RNN-MLP}}

\end{figure}

\begin{figure}

\begin{centering}
\includegraphics[scale=0.4]{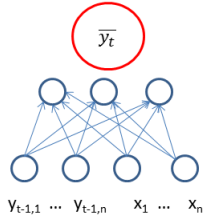}
\par\end{centering}

\caption{CRF-PRCPT - Perceptron architecture including predictions from the
previous time step\label{fig:CRF-PRCPT}}

\end{figure}

We compare and contrast the architectures of different neuroCRFs using
the same methodology:
\begin{enumerate}
\item Combination of two multilayer perceptrons (CRF-MLP) trained with backpropagation
where one MLP learns the weights of the transition factors between
labels and another MLP learns the weights of the observation factors. 
\item Recurrent neural network \cite{elman1990finding} (CRF-RNN) trained
with backpropagation where the activations of the hidden units at
the previous time step are added to the inputs at the next time step
in the sequence. 
\item Structured perceptron \cite{collins2002discriminative} (CRF-PRCPT)
as described above \ref{sec:Related-Work}. 
\end{enumerate}
The different architectures compared are illustrated in Figs. \ref{fig:CRF-MLP},
\ref{fig:RNN-MLP}, and \ref{fig:CRF-PRCPT}. Algorithm \ref{alg:Forward-function}
describes the forward function of the Viterbi algorithm to compute
$P(\bar{y}|\bar{x},\bar{\theta})$ in log space for CRF-MLP. 

\begin{algorithm}
input: model, $\bar{x}$, $\alpha[t-1]$ 

{\footnotesize{}//model, observations at time $t$, and $\alpha[]$}{\footnotesize \par}

{\footnotesize{}//model consists of one MLP for observations, $MLP_{obs},$
}\\
{\footnotesize{}//and one MLP for edges, $MLP_{edges}$. }{\footnotesize \par}

{\footnotesize{}//$\alpha[t-1]$ is a list of tuples \{$to,from,score\}$maximizing }{\footnotesize \par}

{\footnotesize{}//the score for each $to$ label at the previous step
$t-1$}{\footnotesize \par}

output: $\alpha[t]$

\textbf{forward} (model,$\bar{x}$, $\alpha[t-1]$)=

~~~$obspreds$ $\leftarrow$ predict ($\bar{x}$, $MLP_{obs}$)

~~~$edgepreds$ $\leftarrow$$\textrm{Ø}$

~~~$\alpha[t]\leftarrow\textrm{Ø}$

~~~\textbf{foreach} $label\in Y_{model}$, the set of labels for
the model

~~~~~~$edgepreds$ $\leftarrow$$edgepreds$ $\cup$ \{predict
($label$, $MLP_{edges}$)\}

~~~\textbf{foreach} $label\in Y_{model}$

~~~~~~$to',score'$ $\leftarrow$

~~~~~~~~~$arg\max_{\alpha[t-1]}(score+edgepreds[to]_{label}+obspreds_{label})$

~~~~~~$\alpha[t]\leftarrow\alpha[t]\cup\{label,to',score'\}$

\textbf{return} $\alpha[t]$

\caption{Forward function of Viterbi algorithm for CRF-MLP\label{alg:Forward-function}}

\end{algorithm}

Using stochastic gradient descent (SGD), training occurred when the
label, as optimized for the overall sequence by the Viterbi algorithm,
was incorrect. In addition, the weight updates were modulated with
weight elimination regularization \cite{abu2012learning} for the
MLPs:

\[
w_{ij}=w_{ij}-\eta x_{i}(\delta_{j}+2\lambda\frac{w_{ij}}{(1+w_{ij}^{2})^{2}})
\]

where $w_{ij}$ is the weight on the connection between $node_{i}$
and $node_{j}$, $i$ and $j$ denoting different contiguous layers,
$\eta$ is the learning rate, $x_{i}$ is the activation at $node_{i}$,
$\delta_{j}$ is the gradient at $node_{j}$ as calculated by the
backpropagation algorithm, and $\lambda$ is the regularization parameter. 

\begin{figure}
\begin{centering}
\includegraphics[scale=0.29]{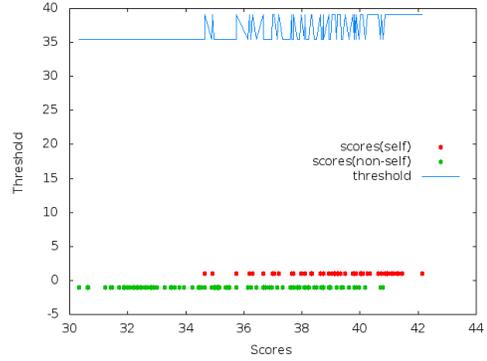}
\par\end{centering}

\caption{Threshold determination (coefficient of regression=37.25, r-square=0.31)\label{fig:Threshold-determination} }

\end{figure}

Our sequence classification metric is based on authentication biometrics
using the false rejection rate (FRR) or false negatives and the false
acceptance rate (FAR) or false positives leading to the identification
of self vs. non-self. Several models are trained, each representing
one type of sequence. Therefore, the conditional probability of a
sequence given a model is not a calibrated probability because the
training data is not assumed to represent the true distribution of
all possible sequences and we do not know about other models (but
we have examples of non-self). However, we can calibrate the score
obtained evaluating sequences of self and non-self against the model
with a threshold to report the equal error rate (EER), where the FRR
equals the FAR, as follows. We map the scores obtained against testing
examples of self to the positive class and scores obtained against
testing examples of non-self to the negative class to obtain a linear
model minimizing the sum of square errors separating the classes.
We take the coefficient of regression for our threshold from which
to compute the FRR and the FAR. Figure \ref{fig:Threshold-determination}
illustrates how the threshold is determined from the scores of examples
of self and non-self. We report results from the evaluation of this
threshold on the validation set itself for a performance approximation.
We report the r-square from the linear model to show the calibration
of scores. The accuracy result is computed from the FRR and the FAR
while the token accuracy is computed as the frequency of correct labels
in the sequences. The F-score combines sequence classification accuracy
and token accuracy results.

\section{Empirical Evaluation \label{sec:Empirical-Evaluation}}

The MLPs used a fixed learning rate set at 0.5 and a regularization
parameter $\lambda$ set at 0.001. All neural nets used 1000 SGD examples
or less if convergence to zero-error occurred during training. The
weights for all neural nets were initialized to small values drawn
from a normal distribution with zero mean and standard deviation 0.00015.
The number of hidden nodes for the MLP-based architectures, CRF-MLP
and CRF-RNN, was set to $\frac{n_{i}+n_{o}}{4}$ as in \cite{goecks2000learning},
where $n_{i}$ is the number of inputs and $n_{o}$ is the number
of outputs. No attempt has been made to optimize the hyper-parameters
of the different learners. The sigmoid function was the activation
function for the nodes in the hidden layer and the derivative of the
square loss function propagated the error at the output nodes. 

The propagated loss in the structured perceptron was the difference
between the predicted outcome and the actual outcome for each feature
function relating inputs to outputs (0/1 loss). In addition, a mini-batch
approach was used where the prediction error was averaged over 5 examples.

\subsection{OCR Dataset}

\begin{figure}
\begin{centering}
\includegraphics[scale=0.29]{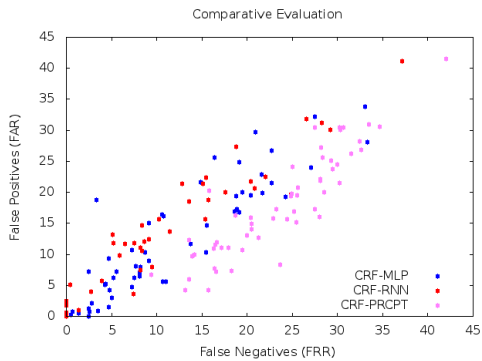}
\par\end{centering}

\caption{Comparative results from neural-network based CRFs on the OCR dataset.
Each data point is the average results for one word over 5 iterations.\label{fig:Comparative-results-OCR}}
\end{figure}

The OCR dataset \cite{kassel1995comparison} contains 52152 16x8 raster
images of letters composing 55 distinct words of length ranging from
3 to 14. The number of examples per word varies from 71 to 151. Table
\ref{tab:OCR-dataset-characteristics} describes this dataset characteristics.
A model is built for each word and tested against other words of the
same length. At each iteration, the data for each word is randomly
partitioned into 2/3 training and 1/3 testing to evaluate the FRR
and compared against a random sample of \textasciitilde{}100 exemplars
of different letters of the same length to evaluate the FAR. Table
\ref{tab:Comparative-results-OCR} and Fig. \ref{fig:Comparative-results-OCR}
illustrates the results averaged over 5 iterations. While there is
no significant statistical difference in the accuracy results between
CRF-MLP and CRF-RNN, there is a significant statistical difference
(two-sided p-value < 0.05) for the accuracy results between CRF-MLP
and CRF-PRCPT and also between CRF-RNN and CRF-PRCPT. However, CRF-PRCPT
has a greater token accuracy value at a significant statistical difference
from both CRF-MLP and CRF-RNN. There is also a greater token accuracy
value at a significant statistical different between CRF-MLP and CRF-RNN. 

\begin{table}
\begin{centering}
\begin{tabular}{|c|c|c|}
\hline 
\textbf{Word} &  & \tabularnewline
\textbf{Length} & \textbf{\#words} & \textbf{\#examples}\tabularnewline
\hline 
3 & 9 & 1283\tabularnewline
\hline 
5 & 4 & 568\tabularnewline
\hline 
6 & 6 & 768\tabularnewline
\hline 
7 & 5 & 695\tabularnewline
\hline 
8 & 6 & 750\tabularnewline
\hline 
9 & 8 & 1047\tabularnewline
\hline 
10 & 5 & 584\tabularnewline
\hline 
11 & 3 & 304\tabularnewline
\hline 
12 & 2 & 298\tabularnewline
\hline 
13 & 3 & 313\tabularnewline
\hline 
14 & 3 & 266\tabularnewline
\hline 
\end{tabular}
\par\end{centering}

\caption{OCR dataset characteristics\label{tab:OCR-dataset-characteristics}}
\end{table}

\begin{table*}[t]
\begin{centering}
\begin{tabular}{|c|c|c|c|c|c|c|}
\hline 
 & \textbf{Avg. } & \textbf{Avg. } & \textbf{Avg. } & \textbf{Avg.} & \textbf{Avg. } & \textbf{Avg.}\tabularnewline
\hline 
\textbf{Methods} & \textbf{FRR(\%)} & \textbf{FAR(\%)} & \textbf{Acc. (\%)} & \textbf{$R^{2}$} & \textbf{Token Acc.(\%)} & \textbf{F-score}\tabularnewline
\hline 
CRF-MLP & 11.81$\pm13.08$ & 12.37$\pm13.89$ & 87.90$\pm08.84$ & 0.65$\pm0.31$ & 93.03$\pm04.36$ & 0.90$\pm4.53$\tabularnewline
\hline 
CRF-RNN & 8.95$\pm14.04$ & 11.01$\pm14.68$ & 90.02$\pm10.18$ & 0.66$\pm0.34$ & 85.17$\pm07.70$ & 0.87$\pm3.73$\tabularnewline
\hline 
CRF-PRCPT & 23.59$\pm09.63$ & 18.06$\pm11.19$ & 79.17$\pm07.38$ & 0.38$\pm0.20$ & 95.44$\pm02.96$ & 0.86$\pm4.95$\tabularnewline
\hline 
\end{tabular}
\par\end{centering}

\caption{Comparative results from neural-network based CRFs on the OCR dataset
averaged over 5 iterations.\label{tab:Comparative-results-OCR}}
\end{table*}

\subsection{Web Analytics}

\begin{figure}
\begin{centering}
\includegraphics[scale=0.29]{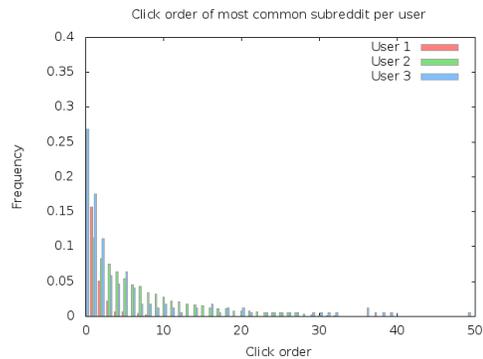}
\par\end{centering}

\caption{Frequencies of the most common subreddit by order visited for 3 users\label{fig:Common_Subreddits}}

\end{figure}

Another type of complex sequence can be found in our online activities.
We extend previous work done in the context of Web browsing \cite{abramson:socialcom}
to profile authentication from social media activities of Reddit users.
Reddit is a public forum where anybody can create \textit{subreddits}
on any topics. We retrieved posts and comments using the Python Reddit
API Wrapper (PRAW) from users associated to a seed user through at
least one comment and from this pool of users selected at random 50
active users. There is a hard limit from Reddit to retrieve only the
last 1000 posts and comments. The session pause delimiter (set to
30 minutes in Web browsing) was extended to one hour due to the sparsity
of posts and/or comments. In contrast to Web browsing, the data of
Reddit activities is sparse as evidenced by the large number of singleton
sessions (Table \ref{tab:Web-dataset-characteristics}). Empirical
analysis shows that the frequencies of the most common subreddits
for a particular user follow the temporal order in which they were
accessed (Fig. \ref{fig:Common_Subreddits}). As stated in \cite{shi2011implicit},
commonality is not a good discriminator. The entropy of the subreddits,
as a measure of commonality, decreases according to their position
in the session. To capture discriminating sequences, we modeled sessions
of Reddit activities as n-gram sequences of length 4 and ignored shorter
sequences. All users have sequences of various length up to length
6. Reddit user activities are modeled with CRFs as a sequence of posts/comments
where the observations are the time-of-day, day-of-week, and subject
header n-grams. A maximum of 100 most common n-grams per subreddit
were extracted from the subject headers for computational efficiency.
The labels are the subreddits of the posts/comments for a strongly
supervised evaluation similar to the OCR dataset. Future work will
evaluate the impact of a weakly-supervised approach with a subreddit
concept hierarchy. Unlike the OCR dataset, the n-gram sequences are
not i.i.d. Each user dataset was temporally partitioned into 90\%
training and 10\% testing.

\begin{table}
\begin{centering}
\begin{tabular}{|c|c|c|c|}
\hline 
\textbf{sequence} & \textbf{Subreddit} &  & \textbf{\#unique}\tabularnewline
\textbf{Length} & \textbf{Entropy} & \textbf{\#sessions} & \textbf{Subreddits}\tabularnewline
\hline 
1 & 9.54 & 18580 & 1307\tabularnewline
\hline 
2 & 8.62 & 5305 & 703\tabularnewline
\hline 
3 & 7.90 & 2146 & 431\tabularnewline
\hline 
4 & 7.37 & 1155 & 293\tabularnewline
\hline 
5 & 6.80 & 655 & 194\tabularnewline
\hline 
6 & 6.49 & 428 & 155\tabularnewline
\hline 
>6 & - & 1231 & 372\tabularnewline
\hline 
\end{tabular}
\par\end{centering}

\caption{Reddit dataset sequence length characteristics for 50 users.  \label{tab:Web-dataset-characteristics}}
\end{table}

\begin{figure}
\begin{centering}
\includegraphics[scale=0.29]{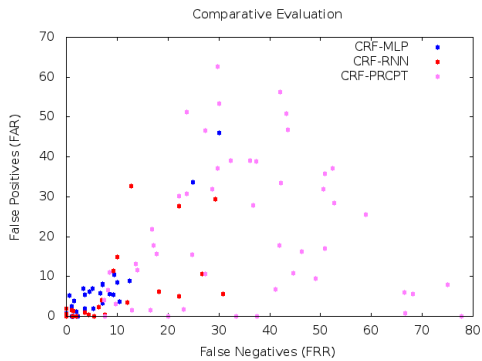}
\par\end{centering}

\caption{Comparative results from neural-network based CRFs on the Reddit dataset.
Each data point is the average results for one individual over 5 iterations.\label{fig:Comparative-results-Reddit}}

\end{figure}

Table \ref{tab:Comparative-results-Reddit} and Fig. \ref{fig:Comparative-results-Reddit}
illustrates the results. There is a significant statistical difference
(two-sided p-value < 0.05) in accuracy between the MLP-based architectures,
CRF-MLP and CRF-RNN, and CRF-PRCPT but no statistical difference in
accuracy between CRF-MLP and CRF-RNN. In addition, there is no statistical
difference in token accuracy between CRF-RNN and CRF-PRCPT. We note
that the correlation between labels is not as stable as in the OCR
dataset which explains why CRF-MLP with an edge prediction network
has lower token accuracy in this dataset. 

\begin{table*}[t]
\begin{centering}
\begin{tabular}{|c|c|c|c|c|c|c|}
\hline 
 & \textbf{Avg. } & \textbf{Avg. } & \textbf{Avg. } & Avg. & \textbf{Avg. } & \textbf{Avg.}\tabularnewline
\hline 
\textbf{Methods} & \textbf{FRR(\%)} & \textbf{FAR(\%)} & \textbf{Acc. (\%)} & \textbf{$R^{2}$} & \textbf{Token Acc.(\%)} & \textbf{F-score}\tabularnewline
\hline 
CRF-MLP & 3.61$\pm10.35$ & 3.83$\pm11.37$ & 96.27$\pm6.99$ & 0.87$\pm0.25$ & 46.89$\pm20.05$ & 0.61$\pm0.11$\tabularnewline
\hline 
CRF-RNN & 4.77$\pm10.90$ & 3.21$\pm09.34$ & 96.00$\pm7.32$ & 0.81$\pm0.26$ & 60.99$\pm19.02$ & 0.72$\pm0.12$\tabularnewline
\hline 
CRF-PRCPT & 34.76$\pm21.37$ & 21.37$\pm21.35$ & 71.92$\pm13.23$ & 0.23$\pm0.25$ & 64.32$\pm16.90$ & 0.66$\pm0.11$\tabularnewline
\hline 
\end{tabular}
\par\end{centering}

\caption{Comparative results from neural-network based CRFs on the Reddit dataset\label{tab:Comparative-results-Reddit}}
\end{table*}

\section{Conclusion \label{sec:Conclusion}}

In summary, we have argued that the discriminative capabilities of
CRFs in the structured prediction task do not necessarily carry over
to the sequence classification task. Toward that end, we have compared
and contrasted different architectures of neuroCRFs in two different
tasks, the OCR hand-writing recognition task and the Web analytics
task, with the same methodology. We have shown that CRFs, as a structured
prediction approach, can also be applied to sequence classification
by training several models for different types of sequence with potentially
different labels, and calibrating the scores of each model (before
softmax squashing) with a threshold determined by a linear model.
While the structured perceptron performs better overall at the structured
prediction task, the discriminative power of MLP based architectures,
CRF-MLP and CRF-RNN, carries over to the sequence classification task
albeit with some degradation in the structured prediction task. We
note that CRF-MLP does better than CRF-RNN in the structured prediction
task of the OCR hand-writing recognition task where the label transitions
are consistent and modeled with an edge prediction network. Future
work will further bridge the gap between sequence labeling and sequence
classification as well as evaluate the impact of a weakly-supervised
approach for user authentication in Web analytics with neural CRFs. 

\bibliographystyle{ieeetr}
\bibliography{backpropcrf}

\end{document}